# Development of a Fast and Robust Gaze Tracking System for Game Applications

Manh Duong Phung, Cong Hoang Quach and Quang Vinh Tran
University of Engineering and Technology
Vietnam National University, Hanoi

*Abstract*— **In this study, a novel eye tracking system using a visual camera is developed to extract human's gaze, and it can be used in modern game machines to bring new and innovative interactive experience to players. Central to the components of the system, is a robust iris-center and eye-corner detection algorithm basing on it the gaze is continuously and adaptively extracted. Evaluation tests were applied to nine people to evaluate the accuracy of the system and the results were 2.50 degrees (view angle) in horizontal direction and 3.07 degrees in vertical direction.**

## I. Introduction

With the rapid development of video game industry, mouse and keyboard inputs are becoming insufficient for the requirements of modern games and players. New and innovative ways of controlling need to be uncovered. In 2003, Sony presented the EyeToy, a camera that is connected to a PlayStation 2 console and tracks the body movements of the players, allowing them to control the on-screen characters by moving their bodies. In 2005, Nintendo presented the Wiimote, a novel gamepad for their console Wii. The Wiimote includes an accelerometer and optical sensor technology that allow games to be controlled by moving the pad in three-dimensional space. The Kinect sensor introduced by Microsoft allowed to not only track the human body for human-computer interaction but also obtain simultaneous both color and depth information for various applications [1][2].

Continuing with the trend of seeking alternative and more intuitive input devices for game interaction, gaze represents a fast and natural input method that can also be exploited [18]. Jonsson compared eye and mouse control as input for two three-dimensional (3D) computer games and found that gaze control was more accurate, game experience was perceived more enjoyable and committing [3]. Smith and Graham studied eye-based input for several game types, with principally 3D navigation. Their results show that participants felt more immersed when using the eye tracker as a gaming input device [4]. Kenny et al. developed a first-person shooter (FPS) game that logs eye tracking data, video data and game internal data simultaneously. They found that players fixate the center of the screen for a majority of the time [5]. These results raise the promising of gaze integration into modern game applications.

In the field of eye tracking, a number of techniques now have been proposed to capture human's gaze, such as electro-oculography (EOG), the photoelectric method, and video-oculography (VOG) [6][7]. Each methodology has its strengths and limitations. The EOG technique measures signal intensity obtained from periphery muscles associated with eye movements. The photoelectric technique tracks the limbus of the eye by measuring the amount of scattered light. Both EOG and photoelectric methods suffer from a number of limitations including unreliably vertical eye gaze measurements, low sensitivity and limited bandwidth [8-10].

In recent years, with the improvements in computing power and image processing, the VOG is becoming more and more attractive, due to its high accuracy, insignificant artifacts and wide applicability [11][20]. At present, most VOG systems use infra-red cameras to get high contract eye images; the user's eyes are illuminated by an infra-red source [20]. This approach although gives results with high accuracy, research shows that the use of infra-red light source may result in potential eye hazards such as eye dryness, lens burning, and retina thermal injury [12][13].

Understanding the limitation of near-infrared camera, several researchers developed gaze tracking algorithms for visual camera. Matsumoto et al. proposed a gaze tracking method that uses a stereo camera to detect iris contour [14]. This method allows the user to move her head freely within the visual field of the stereo camera. The method works well even in daylight conditions. Because it is robust, there are several application areas like gaze detection while driving a car. Kim and Ramakrishna presented an approach to measure the eye gaze via images of the two irises [15]. Their method uses a small mark attached to the glasses stuck between two lenses as a reference point to measure the movement of iris and to eliminate the influence of head displacement. In fact, any additional attachments is resulted in user discomfort and should be avoided.

In this study, we propose a novel algorithm for gaze tracking system using a visual camera. The system does not require user to wear anything on her head, and she can move her head freely. Evaluation tests show that the accuracy of the system is acceptable and sufficient for game applications.

## II. Gaze Tracking Methodology

The proposed system extracts user's gaze through three processes. First, the center of iris is detected from the captured image by a double circle fitting algorithm. The variance projection function (VPF) is then employed to detect the eye corner. The eye-ball model is finally applied

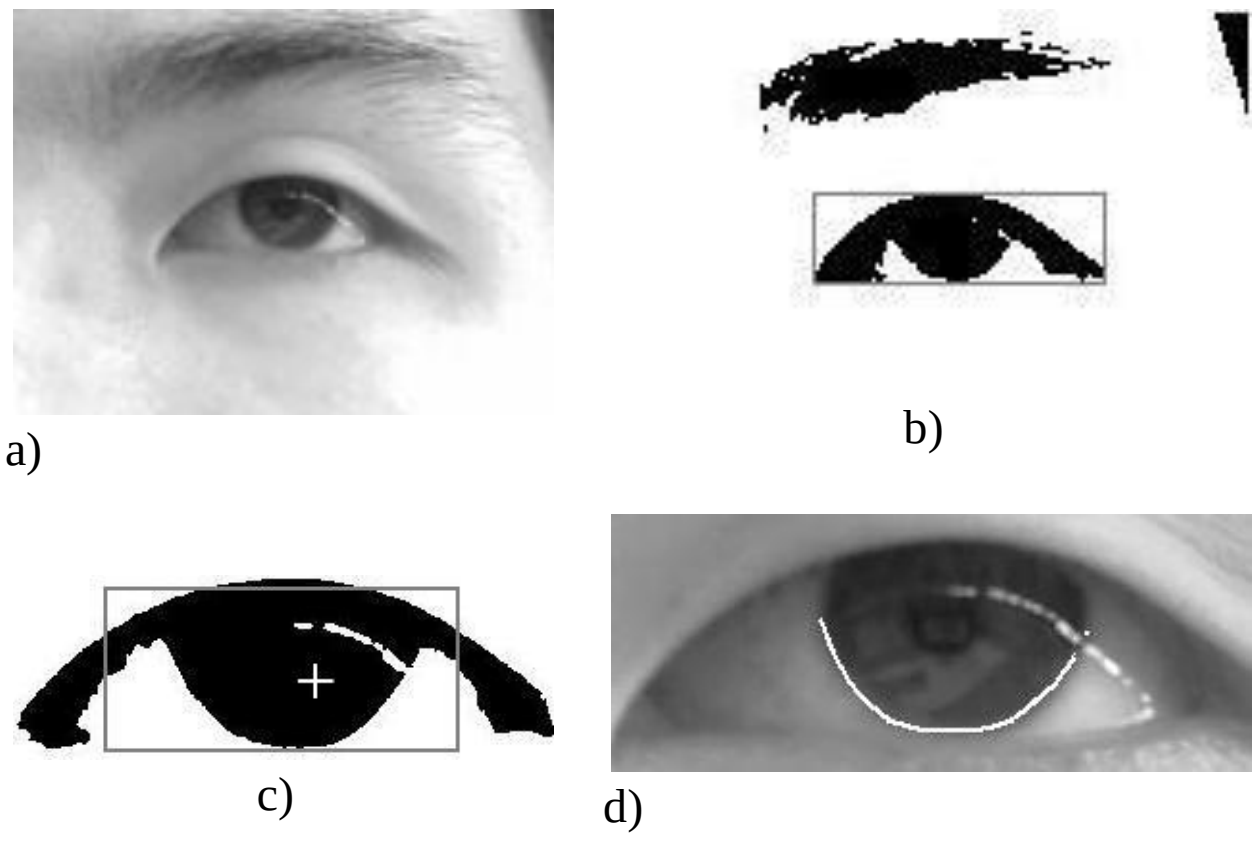

Fig. 1. Sample detection

a) Captured eye image, b) Image segmentation and detected eye area, c) Refined eye area and detected iris position, d) Detected sample points.

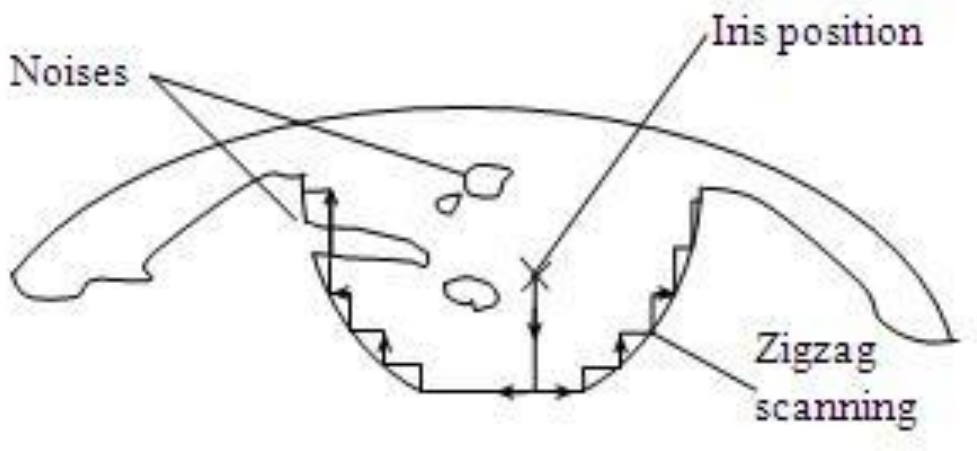

Fig. 2. Detecting sample points by zigzag scanning

to extract the gaze of user.

## A. *Iris Center Detection*

The center of the iris is detected by the logical steps described below.

First, an eye image is captured at a resolution of 640x480 [Fig.1 (a)]. The image is assumed to consist of the eyebrow and eye regions. This simplifies eye detection which is investigated in previous research. To reduce the processing cost, an 80x60 resolution image is generated from the original image. A threshold value is dynamically calculated using isodata algorithm [19]. A segmented image is generated by connecting neighborhood pixels [Fig.1 (b)].

Based on this segmented image, the eyebrow and eye regions are extracted as two largest regions and the eye area is the region below the eyebrow [Fig.1 (b)]. It is then copied to a buffer. All other regions are ignored.

A window is used to detect the iris position by scanning the eye region horizontally. Its height is the height of eye region and its width is 0.15 times the eye region width. These values are set based on the ratio of iris to eye dimensions. During the scanning process, the sum of grey-scale values inside the window is calculated. The position at which the window reaches its maximum value is the position of iris [Fig.1 (c)].

Using the iris position, the eye area is refined and the threshold value is recalculated. This allows the system to work in different lighting conditions.

Now, we return to the original image. All known data included threshold value, eye region and iris position are rescaled. To detect the sample points, points in the iris but at the iris-sclera border, we check the following conditions:

- Starting from the iris position, a vertical scan is made, using the grey-scale threshold, to detect the iris pixel just inside the iris-sclera border.

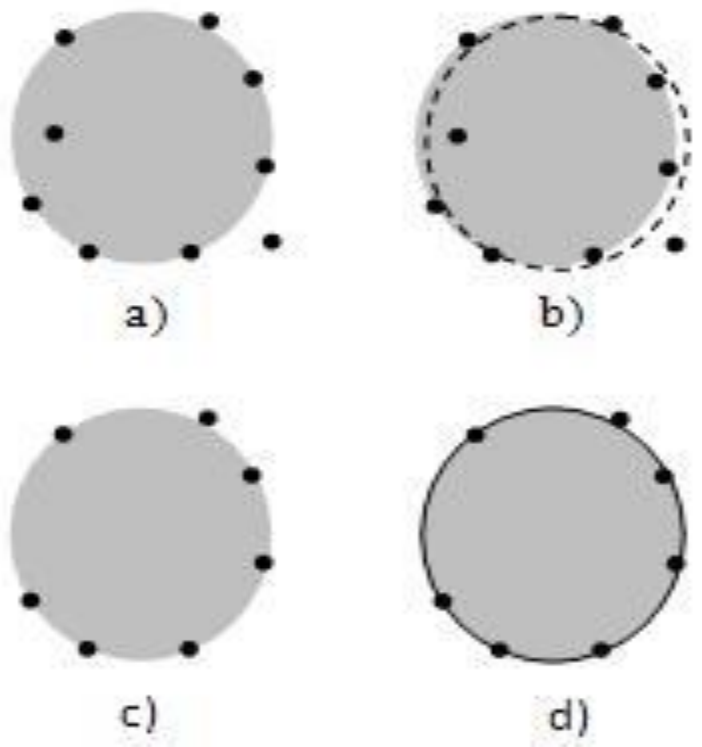

Fig. 3. Double circle fitting

a) Sample points, b) first fit of circle, c) Noise removing, d) Estimated circle.

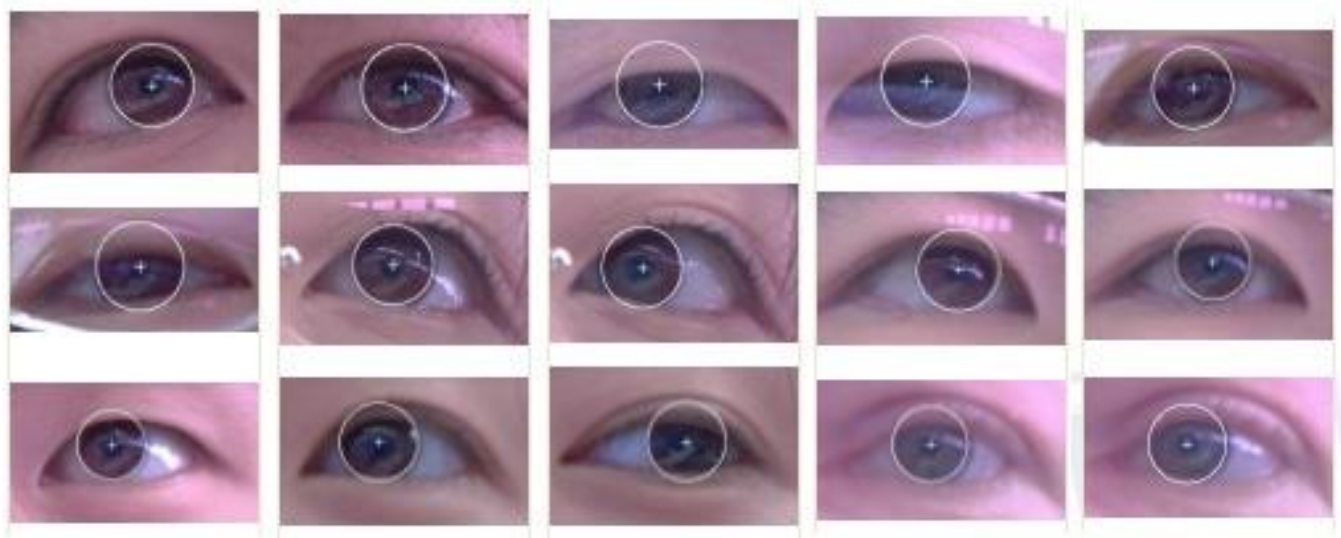

Fig. 4. Iris center detection

- The scan is then extended left and right by raising the scan line height by one pixel and extending it outward until the next sample point is discovered.

With these steps, the system detects sample points by following zigzag paths instead of line by line as conventional. As the result, the detecting process is fast and noise inside or at the edge of the iris is ignored [Fig.2]. Extracted samples are input to the circle fitting algorithm [Fig.1(d)].

The circle fitting algorithm is based on minimizing the mean square distances from the circle to the sample points [21]. Given n points $(x_i, y_i)$, $1 \le i \le n$, the objective function is defined by:

$$F(a,b,R) = \sum_{i=1}^{n} \left(\sqrt{(x_i - a)^2 + (y_i - b)^2} - R\right)^2 \quad (1)$$

where $(a, b)$ is the center of circle and $R$ is its radius. The problem is: determine $(a, b, R)$ to $F$ minimized.

There is no direct algorithm for computing the minimum of $F$, all known algorithms are either iterative (geometric fit) or approximate (algebraic fit) by nature. In this paper, we choose algebraic fit because of its high performance. In algebraic fit, instead of minimizing the sum of squares of the geometric distances, we minimize the sum of squares of algebraic distances. F becomes

$$F(a,b,R) = \sum_{i=1}^{n} (z_i + Bx_i + Cy_i + D)^2 \quad (2)$$

where:

$z_i = x_i^2 + y_i^2, B = \text{-}2a, C = \text{-}2b, D = a^2 + b^2\text{-}R^2$

Differentiating F with respect B, C, D yields a system of linear equations:

$$\begin{bmatrix} M_{xx} & M_{xy} & M_x \\ M_{xy} & M_{yy} & M_y \\ M_x & M_y & n \end{bmatrix} \begin{bmatrix} B \\ C \\ D \end{bmatrix} = \begin{bmatrix} -M_{xz} \\ -M_{yz} \\ -M_z \end{bmatrix} \quad (3)$$

where $M_{xx}$, $M_{xy}$, etc. denote moments, for example $M_{xx} = \sum x_i^2, M_{xy} = \sum x_i y_i$ . Solving this system by Cholesky decomposition gives $B, C, D$ and finally we compute $(a, b, R)$.

As explained in fig.3, after estimating $(a, b, R)$, the distances between sample points and the center of circle are calculated. Sample points which are far from the center are considered noise and are eliminated. Again, circle fitting is performed and the final center is extracted. We call this method *double circle fitting*. Fig.4 shows typical circles estimated by our method.

### B. *Eye Corner Detection*

In eye-based interacting systems, the influence of head displacement on gaze determination is a problem that needs solving. Previous research eliminated this influence by requiring the user to follow an additional process such as calibration or putting a marker on the user's face [11][16]. Any additional process, however, is time consuming and bothers the user. In our system, we use an eye corner as a reference point to automatically offset head movement.

To detect the eye corner, a variance project function (VPF) is employed [17]. This method is based on the observation that some eye landmarks such as eye corners have relatively high contrast which can be detected effectively by the VPF. Suppose $I(x, y)$ is the intensity of a pixel at location $(x, y)$, the variance projection function in vertical direction $\sigma_v^2(x)$ in interval $[y_1, y_2]$ is defined as follow:

$$\sigma_v^2(x) = \frac{1}{y_2 - y_1} \sum_{y_i = y_1}^{y_2} [I(x, y_i) - V_m(x)]^2 \quad (4)$$

where, $V_m(x)$ is mean of the vertical integral projection of $I(x, y)$:

$$V_m(x) = \frac{1}{y_2 - y_1} \int_{y_1}^{y_2} I(x, y) dy \quad (5)$$

Fig.5 presents the VPF along the vertical direction of an synthetic eye. It suggests the following eye corner detection approach.

First, an eye corner area is determined based on the center and radius of iris [Fig.6(a)]. The vertical variance projection function $\sigma_v^2(x)$ and its derivative are then applied to the area to detect the vertical position of the eye corner [Fig.6(b)]. A Sobel edge detector is used to detect the eyelid. The horizontal position is the intersection between the vertical position and the eyelid [Fig.6(c)].

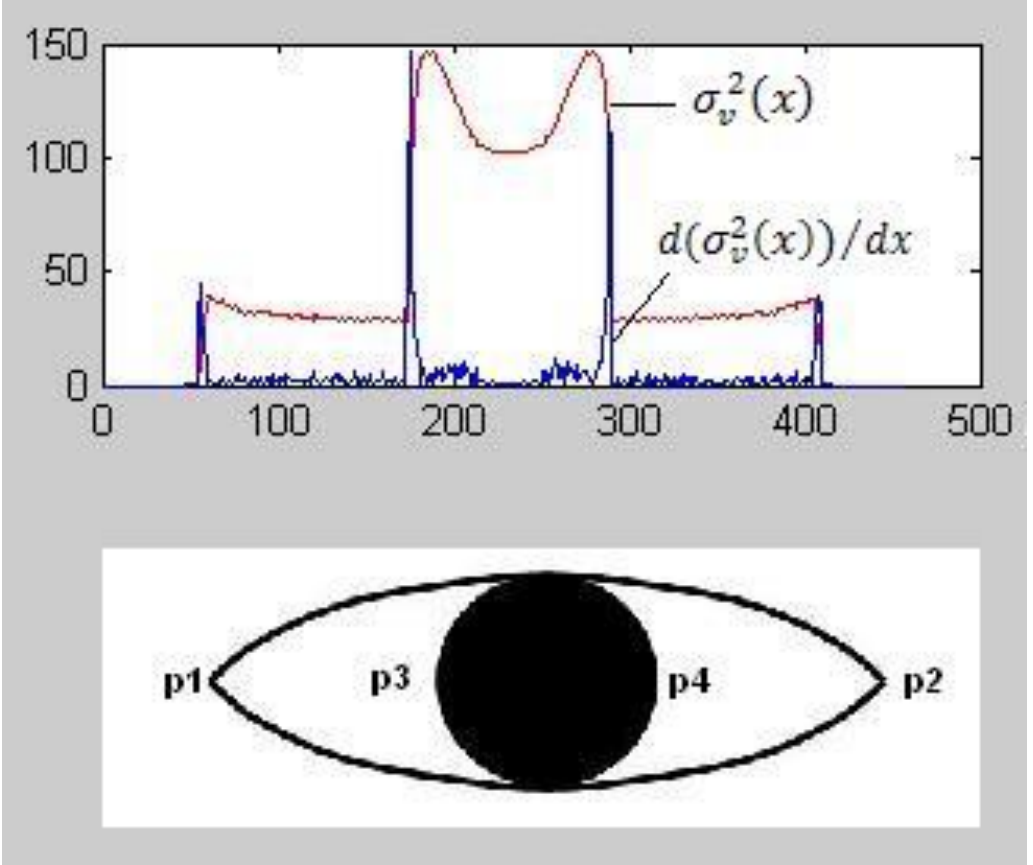

Fig. 5. A synthetic image and its VPF along the vertical direction

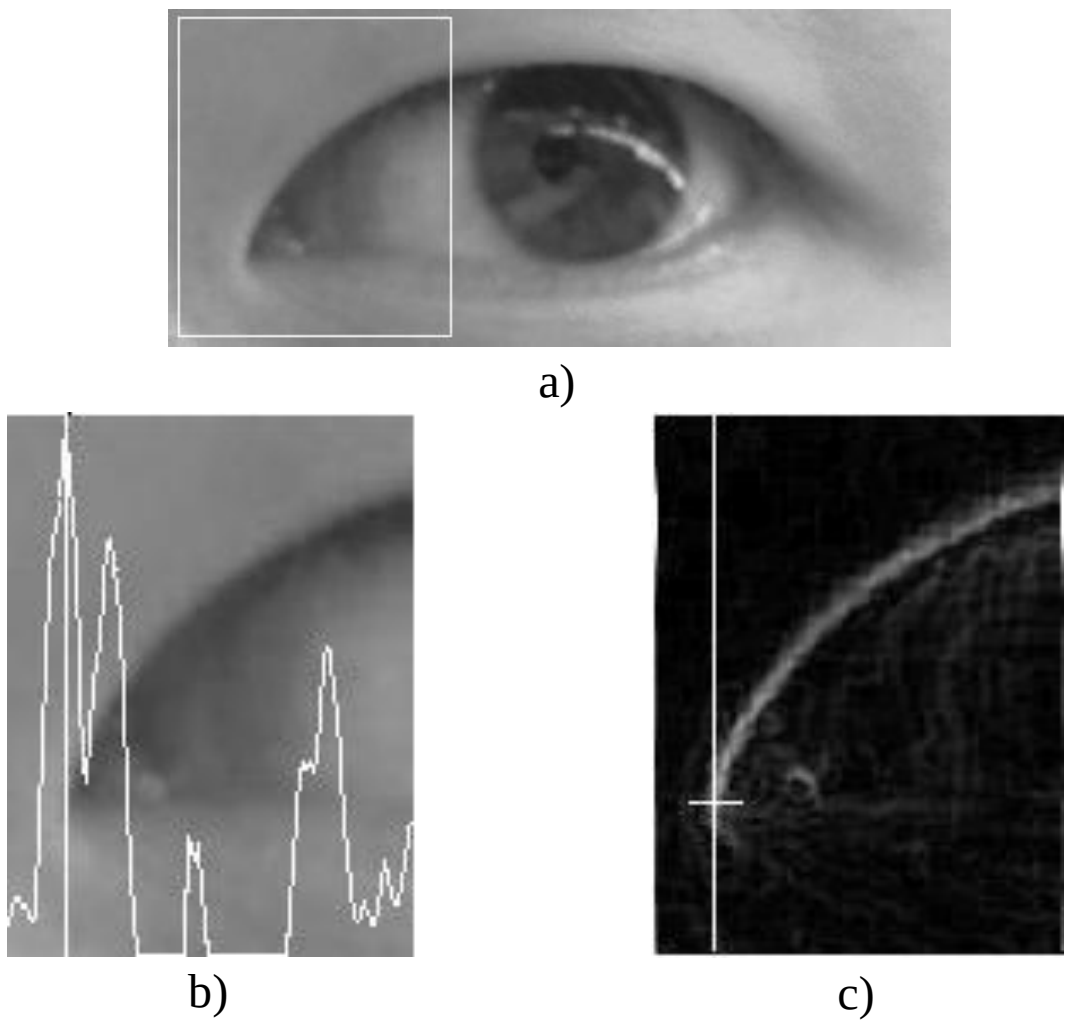

Fig. 6. Eye corner detection
a) Eye corner area, b) Derivative of VPF, c) Eyelid and detected eye corner

### C. *Gaze Estimation*

Information provided by previous steps before gaze estimation consists of:

- The position of eye corner. This involves the position of the face.
- The radius of iris. This is useful in estimating the distance from the camera to the subject.
- The vector from the eye corner to the iris center. This carries information on eye movement.

Based on these data, fig.7 shows the eye-ball model to estimate the gaze [15]. The circle shows the projection of the spherical eyeball. Three gazes (gaze 1, gaze 2, and gaze r) and projection of each iris center and gaze are shown in the figure. Gaze r is the reference gaze. d is the distance from the eyeball surface to the screen plane. d is assumed given or can be intentionally set to fixed value. $r_{ball}$ is the radius of the eyeball, which ranges from 12mm to 13mm (according to the anthropometric data [15]). $\Delta_1$ and $\Delta_2$, measured in section 2, are the displacements of the iris center of gaze 1 and gaze 2, respectively, from that of gaze r in the projection. $g_1$ and $g_2$ are the displacements of gaze 1 and gaze 2 respectively, from gaze r. The input and output are $\Delta_1$, $\Delta_2$ and $g_1$, $g_2$ respectively.

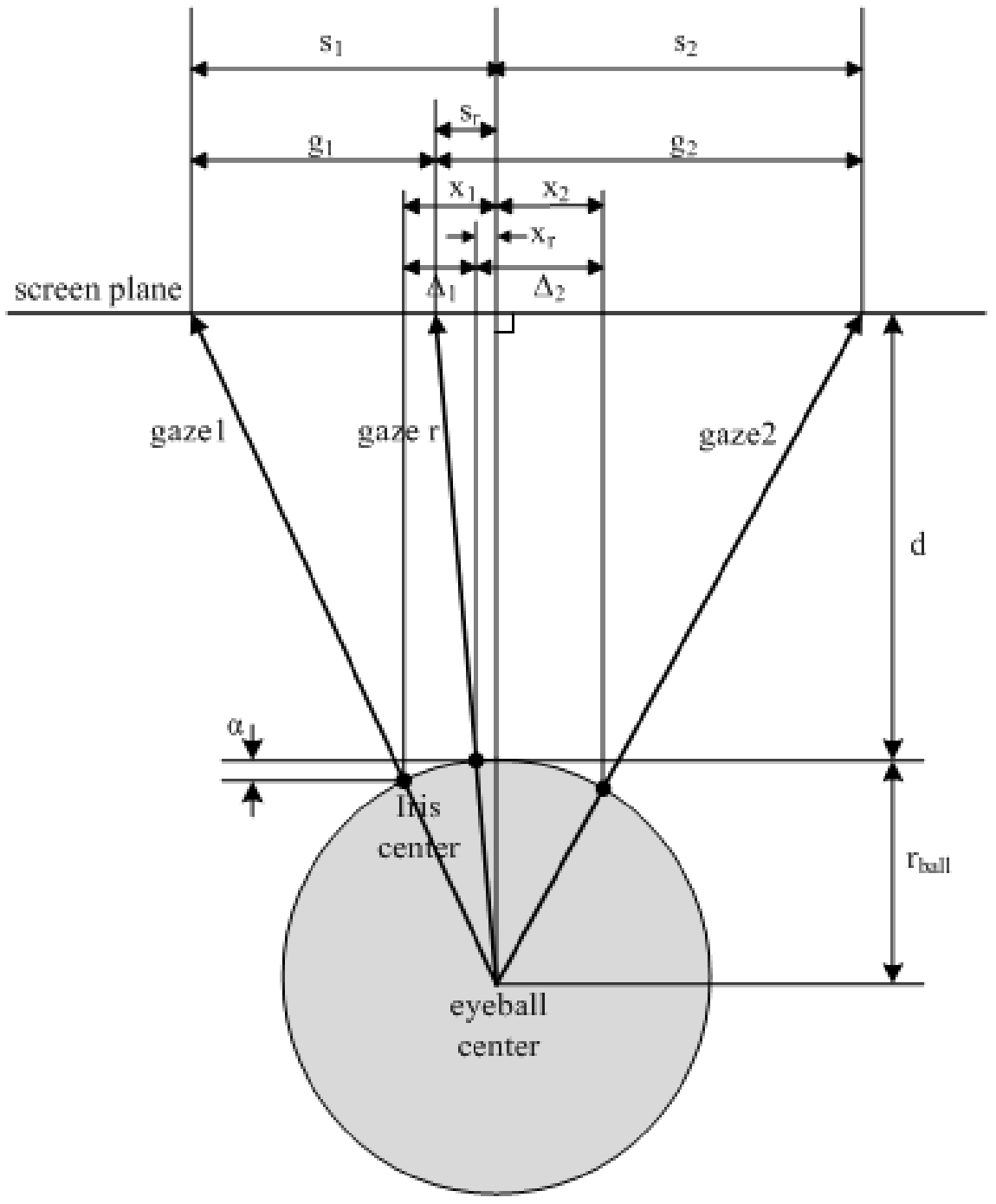

Fig. 7. Geometry model for gaze estimation

If $\alpha = r_{ball} - \sqrt{r_{ball}^2 - x^2} = 0$, then

$$g_1 = \frac{d + r_{ball}}{r_{ball}}(x_1 - x_r) = \frac{d + r_{ball}}{r_{ball}}\Delta_1 \qquad (6)$$

$$g_2 = \frac{d + r_{ball}}{r_{ball}}(x_2 + x_r) = \frac{d + r_{ball}}{r_{ball}}\Delta_2 \qquad (7)$$

In case of head displacement, fig.8 shows the model to estimate the gaze. Based on the eye corner position, we are able to measure the displacement $\Delta_{ref}$ of head which corresponds to the displacement of eye corner. If g is the distance between two screen points, using equations 6 and 7, we get:

$$g = \frac{d + r_{ball}}{r_{ball}}(\Delta_2 - \Delta_1) + \Delta_{ref} \qquad (8)$$

### *D. Calibration Process*

According to previous section, one reference point is necessary to estimate the displacement of gaze on the screen. In most cases, the estimated gaze position, however, contains measurement error due to the error of parameters such as d, $r_{ball}$, $\Delta_{ref}$... To reduce this error, additional calibration process is necessary. In our system, the linear approximation is used and equation 8 becomes:

$$g = k\left[\frac{d + r_{ball}}{r_{ball}}(\Delta_2 - \Delta_1) + \Delta_{ref}\right] \qquad (9)$$

To determine k, one more calibration point needs to be added resulting in 2 points for the calibration process.

## III. EXPERIMENTS

### *A. Experimental implementation*

We implemented an eye tracking system that detects gaze in real time. An overview of the system is shown in

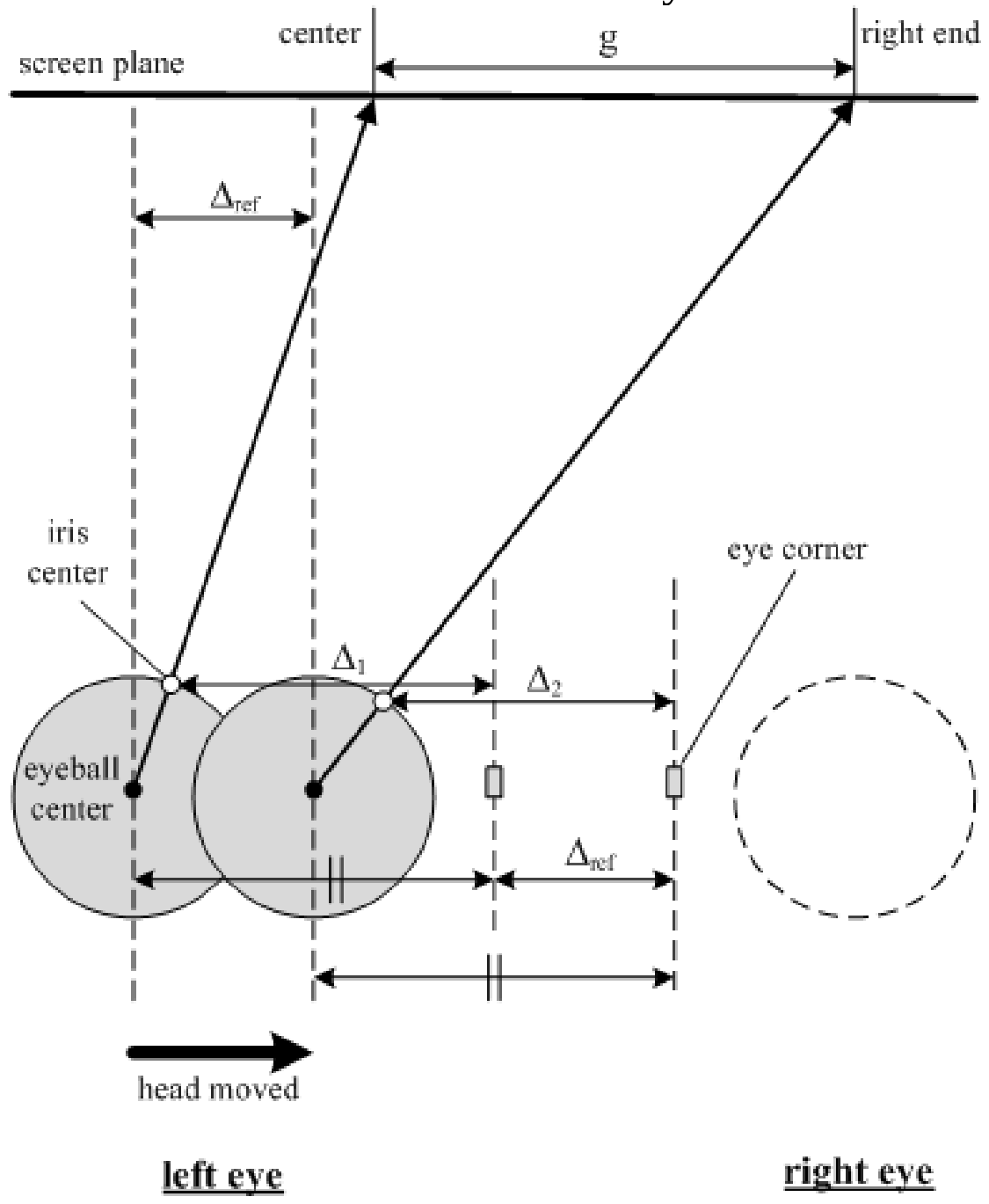

Fig. 8. Geometry model with a head displacement

fig.9. It consists of a DSP board with power supply, a Sony camera and a display. There is no computer in the system. The DSP board is TMS320C6416 DSP Starter Kit of Texas Instrument with 1 GHz processor, 512 words of Flash and 16MB SDRAM. The captured eye image has a resolution of 640x480 pixels and the sampling rate is 30 fps (frame per second).

TABLE 1
EXPERIMENT RESULTS

| Points | Original Position | | Gaze Position Error | |
|---|---|---|---|---|
| | x (mm) | y (mm) | Δx (mm) | Δy (mm) |
| 3 | 0 | 315 | 28 | 28 |
| 4 | 280 | 315 | 32 | 24 |
| 5 | 0 | 158 | 30 | 37 |
| 6 | 280 | 158 | 21 | 33 |
| 7 | 560 | 158 | 25 | 39 |
| 8 | 280 | 0 | 28 | 43 |
| 9 | 560 | 0 | 35 | 40 |

We conducted the experiment as follow:

**Setup** An additional display (700mmx394mm) and a sound player were added to the system. The function of the display is to pose experimental points to subjects [Fig. 10]. The sound player plays sounds to guide subjects.

**Subjects** Nine subjects participated the experiment. Six subjects had naked eyes, and the rest wore eyeglasses.

**Procedure** The system was calibrated by the two-point calibration method. During the calibration, subjects were asked to look at calibration markers on the display, one after another, controlled by sound signals.

When the calibration was finished, seven points appeared at different positions, one after the other, and for each point, the subjects were asked to uninterruptedly look at until a sound was played.

**Result** Table.1 shows the average measurement error in terms of x-coordinate, y-coordinate for subjects. The distance between subjects and the screen was slightly different in each session. If we assume that the distance between the screen and the user is 650mm, which is the typical distance in the implemented system, 50mm on the screen equals a view angle of 4.40 degrees. In horizontal direction, the measurement error was between 21mm and 35mm, the average was 28mm equivalent to a view angle of 2.50 degrees. In vertical direction, the measurement error was increased; it was between 24mm and 43mm, the average was 35mm equivalent to a view angle of 3.07 degrees. In the best case, the error was 21mm, which is equivalent to a view angle of 1.85 degrees. In both cases, the x-coordinate accuracy was better than y-coordinate accuracy. This result matches the fact that eyelids cover upper and lower parts of our eyes.

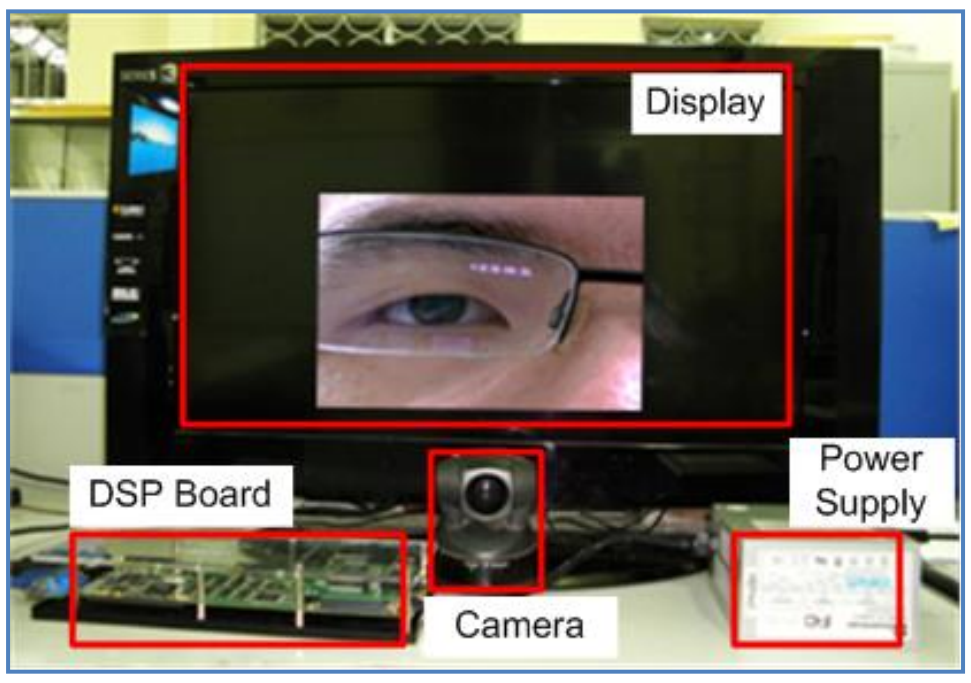

Fig. 9. Overview of the gaze tracking system .

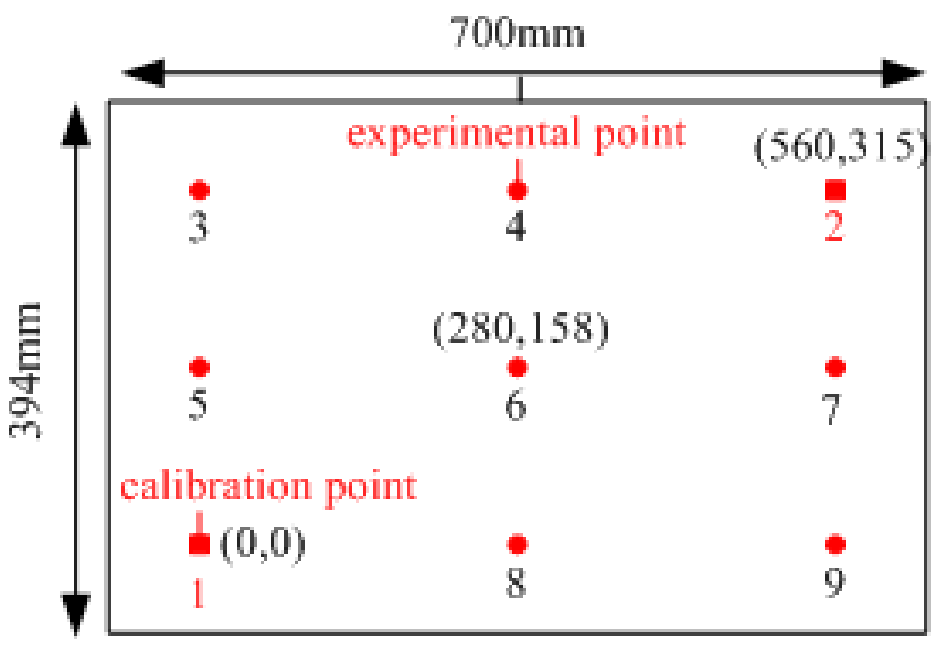

Fig. 10. Experimental implementation

In addition, we also estimated the computational expense of the algorithm by repeating it for each image frame until the system is suspended. At 30 fps (frames per second), the system worked as usual when the algorithm was applied 10 times to each image frame. This implies the maximum sampling rate of the system is 300Hz.

During the experiment, the system sometimes failed to detect the iris center. The reason is that a large part of the iris is hidden which resulted in insufficient number of sample points [Fig.11]. It is difficult to detect iris centers in these situations.

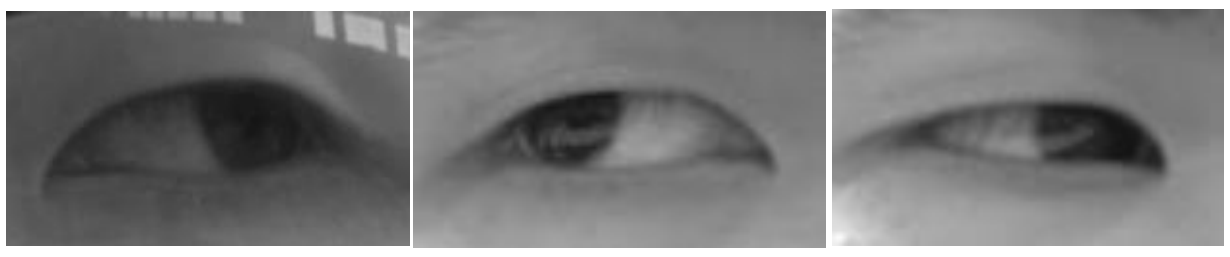
Figure 11. Failed detection situations

### B. *Demonstrations*

To demonstrate the applicability of proposed approach, we implemented a game named "Hammer Heads". This game is simulated from a popular game for in which each player is equipped with a hammer to beat monsters emerged randomly from holes. In our implementation, players use their eyes, instead of hands, to shoot monsters appear randomly on the screen. Fig.12 shows the interface of the game. Nine subjects (three females, six males) were chosen to experience the system.

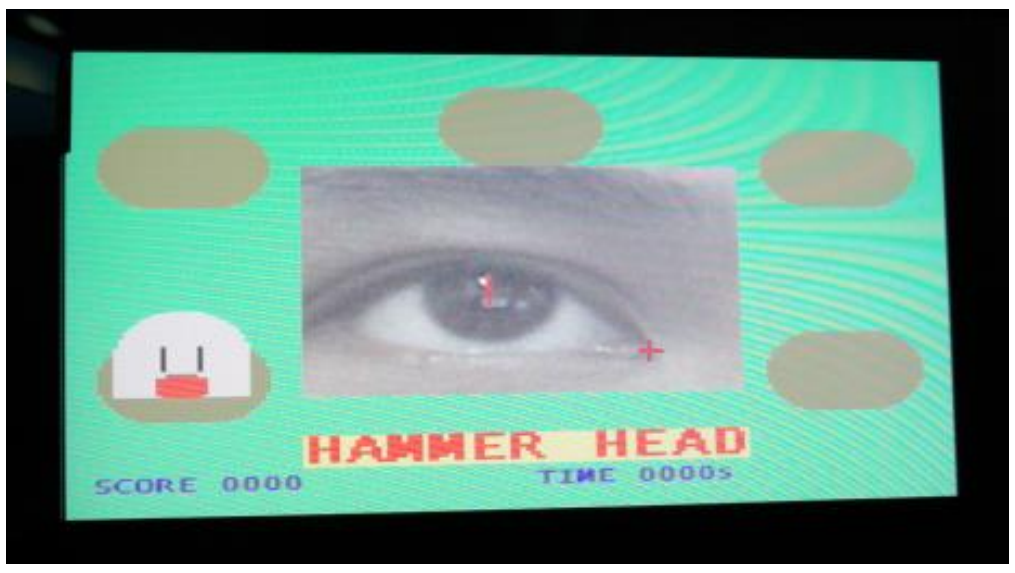

Fig. 12. Game Hammer Heads

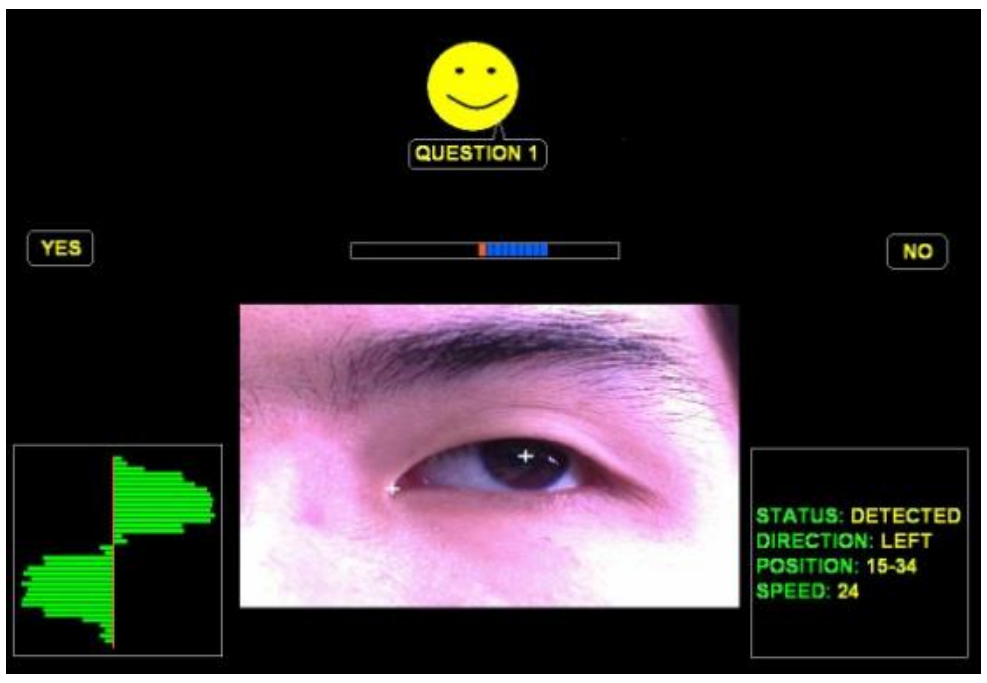

Fig. 13. Survey Machine

To initialize, players are first looked at two crosses appeared at bottom left and top right of the screen. This step is necessary for the calibration process and is a sign to start the game. Players are then use their eye to shoot monsters appearing randomly on the screen. The more monsters a player shoots in a certain amount of time, the more points he scores.

During the interacting process, eye detection was sometimes failed and players had to try again. The reason is that players' hair covered the eyebrow result in wrong detection of eye area which was not focused by our system. After the demonstration, all players enjoyed this new interaction and asked about the system.

In another application, we developed a survey machine in which users use their eyes to answer the survey questions by gazing at Yes or No buttons on the screen [Fig.13]. Since there are only two answer options, the calibration process is reduced to one point located at the center of screen. The initialization process is therefore naturally started when user gazes at questioner for 3 seconds. Nice subjects were succeeded in interacting with the machine without difficulties.

In fact, applications of the system are not limited to games; the many application areas are possible, including an online document browser, device controlling, and information board.

## IV. Conclusion

An easy-setup gaze tracking system was developed for game applications. A novel approach was proposed to estimate the user's gaze from the iris center, eye corner and eye-ball model. Evaluation tests confirmed that the accuracy of gaze detection is about a view angle of 3.07 degrees. We are now trying to improve the accuracy. Development of a mechanism that can automatically determine the distance between user and screen, and control a camera to focus on user's eye is our future work.